\documentclass[11pt,a4paper]{article}
\usepackage[hyperref]{emnlp2020}
\usepackage{times}
\usepackage{latexsym}

\usepackage{amsfonts}
\usepackage{amsmath}
\usepackage{float}
\usepackage{graphicx}
\usepackage{hyperref}
\usepackage{subcaption}
\usepackage{stfloats}
\usepackage{xcolor}

\graphicspath{{images/}}

\usepackage{microtype}

\aclfinalcopy 

\newcommand{\startsquarepar}{
    \par\begingroup \parfillskip 0pt \relax}
\newcommand{\stopsquarepar}{
    \par\endgroup}

\title{Generalizing Emergent Communication}

\author{Thomas A. Unger         \\
  University of Amsterdam       \\
  Amsterdam, The Netherlands    \\
  \texttt{thomas.a.unger@icloud.com}    \\\And
  Elia Bruni                    \\
  Pompeu Fabra University       \\
  Barcelona, Catalonia          \\
  \texttt{elia.bruni@gmail.com} \\}

\date{}

\begin{document}
\maketitle
\begin{abstract}
We converted the recently developed BabyAI grid world platform to a sender/receiver setup in order to test the hypothesis that established deep reinforcement learning techniques are sufficient to incentivize the emergence of a grounded discrete communication protocol between generalized agents. This is in contrast to previous experiments that employed straight-through estimation or specialized inductive biases. Our results show that these can indeed be avoided, by instead providing proper environmental incentives. Moreover, they show that a longer interval between communications incentivized more abstract semantics. In some cases, the communicating agents adapted to new environments more quickly than a monolithic agent, showcasing the potential of emergent communication for transfer learning and generalization in general.
\end{abstract}

\section{Introduction}
The traditional approach to language modeling suffers a fundamental flaw. While models such as GPT-2 \citep{radford2019language} show that capturing the statistics of natural language corpora can lead to impressive results, they are ultimately divorced from the reality they are supposed to describe; they capture the statistics of descriptions of things, rather than the statistics of the things themselves. This approach limits the grounding of semantics.

Some recent works have attempted to improve upon this by instead employing communication games, in which agents are incentivized to use communication as a tool in order to optimize for some objective. This frames communication as a means toward an end, instead of a thing to merely mimic. The corresponding learning process, sometimes called \textit{language emergence}, can then be framed as a negotiation process through which agents establish a common semantics. 

Besides the fact that such a treatment of communication can be conducive to cooperation, there are a number of other benefits to this approach. Since the agents must transmit relevant information to complete a given task, it naturally coaxes agents into explaining themselves, so that we can listen in and perhaps understand what they are thinking. Another benefit is its intrinsic interactivity, meaning that we can position ourselves as one of the agents, to give commands, to ask for clarification, or vice versa.

Lastly, framing communication as a tool is in parallel with the process by which natural language emerged and is maintained. Any natural language which is actively in use is a living language; it is ever-changing and in effect ever-emerging. For example, many humans adopt and invent jargon in their professions; it is arguably desirable that artificial agents that have been primed to use natural language will nonetheless similarly and independently develop artificial jargon to aid in the pursuit of their goals.

\subsection{Related Work}
Typical methods to train communicating agents are those under the umbrella of multi-agent reinforcement learning (MARL). With the advent of deep learning, these further tend to fall under multi-agent deep reinforcement learning (MDRL). \newcite{DBLP:journals/corr/abs-1810-05587} gave a brief survey of recent works using MDRL. They subdivided these into four categories. Namely, the analysis of emergent behaviors, learning communication, learning cooperation and agents modeling agents. They further stated non-stationarity, the increase of dimensionality and the credit-assignment problem as specific difficulties in the use of MDRL methods.

One difficulty specific to the use of discrete symbols, which has made it unpopular in the past, is that their non-continuous nature makes them non-differentiable. This means that communication cannot be optimized in a straightforward way using backpropagation. In order to deal with non-differentiability, \newcite{DBLP:journals/corr/HavrylovT17} employed a straight-through Gumbel-softmax estimator \citep{DBLP:journals/corr/BengioLC13, DBLP:journals/corr/MaddisonMT16, jang2016categorical} in order to study the emergence of discrete communication in simple referential games. \newcite{mordatch2018emergence} similarly proposed the use of a straight-through Gumbel-softmax estimator in a more complex environment. However, the use of straight-through estimation means that the agents have access to each other's internal states, from which they can extract counterfactual information, i.e., ``how would the other agent have reacted if I had said something slightly different?''. This does not generalize to training communication with opaque agents, whose internal states are inaccessible, e.g., humans or hard coded agents.

\newcite{foerster2016learning} and \newcite{DBLP:journals/corr/abs-1810-11187} employed centralized learning, in which agents share parameters that are optimized in a centralized manner. This essentially improves coordination by letting agents talk to copies of themselves, which does not generalize to genuinely separate agents. The latter work also employed continuous communication, which is unlike the discrete nature of natural language.

\newcite{DBLP:journals/corr/abs-1810-08647} criticized the use of centralized learning by \newcite{foerster2016learning} as being ``unrealistic'' and instead resorted to an explicit \textit{model of other agents} (MOA) in order to incentivize the emergence of discrete communication between separate agents in grid worlds. This MOA makes explicit predictions of actions of other agents, which incentivizes the maximization of the mutual information between the messages and the resulting actions. However, hard coding such a mechanism into the agent model makes it inflexible, which renders communicating agents more transparent in the sense that they can make assumptions about each other's inner workings. While this may hasten learning in some cases, it may slow or even prevent learning communication with generalized agents, for which these assumptions do not hold.

\newcite{choi2018compositional} studied the emergence of discrete communication in simple referential games by employing an inductive bias in the agent model in the form of an \textit{obverter} method. This method optimizes the sender's own understanding of messages as a proxy for optimizing the receiver's understanding, relying on the assumption that these correlate. Again, such an assumption may be unwarranted for generalized agents. \newcite{DBLP:journals/corr/abs-1804-03984} did very similar work without employing this mechanism, which however still remained confined to simple referential games.

\section{Method}
In attempting to improve on the work discussed above, we hypothesized that established deep reinforcement learning methods are sufficient to incentivize the emergence of a grounded discrete communication protocol, without relying on unrealistic straight-through estimation methods or specialized inductive biases and instead providing proper environmental incentives. Moreover, in contrast to the simple referential games typically used previously, we tested this hypothesis on the more complex BabyAI platform.

\subsection{BabyAI}
BabyAI \cite{DBLP:journals/corr/abs-1810-08272} is a research platform originally intended to support investigations towards including humans in the loop for grounded language learning. A task in a BabyAI environment consists of a single agent navigating a rectangular grid world to achieve a given objective. Besides being occupied by the agent, a cell may be occupied by an object, which can either serve a purpose toward the objective, or exist purely as a distraction to this objective. The objective of the agent is provided in the form of an instruction in a synthetic language which is a proper subset of English. Following such an instruction may involve learning multiple tasks such as moving, changing orientation and picking up objects. By default, the agent has a partial, egocentric view of the environment.

The entirety of the BabyAI platform comprises an extensible suite of nineteen environments of varying difficulty. This framework promotes \textit{curriculum learning}, in which agents are trained to perform simpler tasks before being trained to perform more complex tasks. Increasingly complex tasks are composed of increasing numbers of subtasks, e.g., picking up a key needed to open the corresponding door.

The authors included various baselines using both imitation learning (IL) and RL training procedures. They found that agents benefited only modestly from pretraining on relevant subtasks.

\subsection{Sender/Receiver Setup}
The BabyAI platform does not natively support a multi-agent setup. Therefore, in order to test our hypothesis, we converted it to a sender/receiver setup, which is publicly available\footnote{\href{https://github.com/thomasaunger/babyai_sr}{\texttt{github.com/thomasaunger/babyai\_sr}}}. A sender/receiver setup is one in which one agent can transmit information to another agent to improve its performance on a given task. This is the typical setup used in simple referential games, but adopting the BabyAI platform complicates the task by introducing dependencies across successive actions.

\begin{figure}[b]
  \centering
  \includegraphics[height=0.85333 in,trim={0 0 0 0},clip]{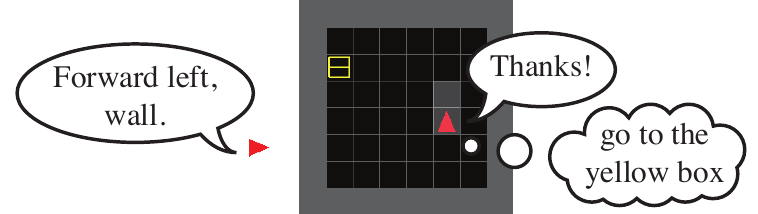}
  \caption{a cartoon depicting our sender/receiver setup in BabyAI. The triangles indicate the two agents; the receiver is inside the environment with its limited, egocentric view indicated by the highlighted cells, while the sender is outside the environment with a global view. The sender must use the available messages and their constituent symbols to transmit any information to the receiver, prompting it to give feedback indicating the utility of the messages. Only the receiver knows the explicit instruction, as indicated by its thought bubble.}
  \label{fig:cartoon}
\end{figure}

There are five incentives that characterize our basic setup, subdivided into two primary and three secondary incentives. Figure \ref{fig:cartoon} summarizes these incentives.

\subsubsection{Primary Incentives}
Each of the two primary incentives provides one of the agents with the basic incentive to actually use the communication channel.

\paragraph{Receiver's Incentive}
There is a differential between the observations of the sender and the receiver. The receiver takes on the role of the single agent in the original BabyAI setup, but in contrast to the original agent has its observations limited to only two cells, namely the cell it is occupying and the cell it is facing. The sender on the other hand views the environment from an ``Archimedean point'', i.e., it has a global view of the environment and can affect the environment only indirectly, namely through any messages it sends to the receiver. This informational differential provides the receiver with an incentive to listen to the sender.

\paragraph{Sender's Incentive}
While the receiver obtains its reward signal directly from the environment, the receiver in turn provides the reward signal to the sender. Section \ref{senders_fee} provides further details. This reward provides the sender with an incentive to actually transmit information relevant to the receiver.

\subsubsection{Secondary Incentives}
While the primary incentives provide the agents with the basic incentives to use the communication channel, the secondary incentives incentivize more precisely how this communication channel is to be used.

\paragraph{Inter-Message Economic Incentive}
The sender sends a single message every $n$ time steps, where $n$ is dubbed the \textit{communication interval}. For example, if $n = 1$, then the sender could get away with dictating one action per message. However, as this value increases, so does the incentive to compile information about multiple time steps into each message, i.e., to use messages economically.

\paragraph{Intra-Message Economic Incentive}
Each message consists of a series of discrete symbols. These messages have a fixed length of $k$ symbols and a fixed symbol vocabulary $\mathcal{V}$. We indicate both the length of the message $k$ and the size of the vocabulary $|\mathcal{V}|$ using the notation $|\mathcal{M}| = |\mathcal{V}|^k$, where $|\mathcal{M}|$ denotes the size of the set $\mathcal{M}$, which is the set of possible messages. Lower values of $k$ and $|\mathcal{V}|$ increase the incentive to compress information within each message, i.e., to use symbols economically.

\paragraph{Encyclopedic Incentive}
The instruction is shown only to the receiver. When there is only one object, the goal is implicit in the sender's observations. However, when it is ambiguous which object is the goal, this information differential provides the sender with an incentive to offer a description of the environment that is more comprehensive than simply the location of the goal, or a list of actions necessary to complete the task.

\subsection{Agent Model} \label{agent_model}
The authors of BabyAI implemented two Advantage Actor--Critic (A2C) agent models to produce baseline RL performances. They referred to these two models as the Large model and the Small model. The latter formed the basis for our own implementation. The small baseline model uses a unidirectional GRU \citep{cho2014learning} to encode the instruction and a convolutional neural network (CNN) with two batch-normalized \citep{ioffe2015batch} FiLM \citep{perez2018film} layers to process both the observation and the encoded instruction. This serves as input for an LSTM \citep{hochreiter1997long}, which integrates it with inputs from previous time steps.

To facilitate communication, we enhanced the agent model with a decoder module and an encoder module. The decoder module consists of an LSTM module whose input is the hidden state of the agent's memory and whose output is a series of vectors, each of which is fed through a linear layer that produces the logits of a probability distribution over $|\mathcal{V}|$ possible symbols. These logits are fed through a softmax layer to produce the actual probability distribution, similarly to how the existing policy module produces a probability distribution over actions. From the resulting sequence of probability distributions are sampled one-hot vectors, constituting a sequence of discrete symbols.

The encoder module likewise consists of an LSTM, whose input is such a sequence of one-hot vectors and whose output is its final hidden state, which is concatenated with the instruction embedding.

\subsection{Training Procedure} \label{training_procedure}
In order to train the agents, we adapted the implementation of the proximal policy optimization (PPO) \citep{schulman2017proximal} algorithm that was used by the BabyAI authors to produce their RL baselines. This is a highly parallelized implementation written in the Python programming language using the machine learning library PyTorch, that can also utilize any graphics processing unit (GPU) supporting CUDA, which we made extensive use of to maximize performance.

The implementation of PPO has a plethora of hyperparameters, such as learning rate, epochs per batch and batch size. For most hyperparameters, changing their default value typically did not increase performance and in fact usually decreased performance. For this reason, we assumed the default values to have been sufficiently tuned for our purposes and thus we refrained from deviating significantly from these values.

\subsubsection{Frame Cadence}
At each time step $t$ in an environment, the receiver is assigned a frame $r_t$, during which it makes an observation, updates its memory and then performs one action. Our implementation of the sender/receiver setup inserts an extra frame $s_i$ assigned to the sender before the frame $r_i$ at every time step $i$ at which the sender emits a message. During these frames, the sender makes an observation, updates its memory and then emits a message $m_i$. This message is placed into a buffer, which is read by the receiver at every time step.

\subsubsection{Sender's Fee} \label{senders_fee}
A2C methods use the policy gradient to optimize an agent's ``actor'' $\pi$, such that actions which lead to a state $s$ with a high value $v_\pi(s)$ are reinforced. These state values in turn are estimated by the agent's ``critic'' $V(s)$. By exploiting the Bellman equation, $V(s)$ can be optimized using \textit{temporal-difference} (TD) learning, i.e., by \textit{bootstrapping} the state-value estimate $V(S_t)$ with the state-value estimate $V(S_{t+1})$, as follows.
\begin{align}
  V(S_t)   &= R_{t+1} + \gamma V(S_{t+1}) \label{eq:1}
\end{align}

While the reward for the receiver is straightforward---the reward as given by the environment---the same is not true for the reward of the sender. We essentially bootstrap the sender's state-value estimate $V_s(S_i)$ with the receiver's state-value estimate $V_r(S_i | m_i)$, where $m_i$ is the message that the sender sent to the receiver at time step $i$. Using equation \ref{eq:1}, we can write $V_s(S_i)$ and $V_r(S_t)$ for the sender and the receiver, respectively, as follows.
\begin{align}
    &\begin{aligned}
        V_s(S_i) &= \color{white} R_{i+1} + \color{black} \gamma V_r(S_i | m_i) \label{eq:4} \\
    \end{aligned}
    \\[\abovedisplayskip]
    &\begin{aligned}
        V_r(S_t) &= R_{t+1} + \gamma V_r(S_{t+1}) \\
    \end{aligned}
\end{align}
As visible in equation \ref{eq:4}, the sender does not bootstrap using its own value estimates $V_s(S_{i+1})$ at all, but rather treats every message emission as a separate task. That is, from the sender's perspective, every ``episode'' lasts a single time step and the sender experiences a number of such ``episodes'' equal to the number of messages emitted during what is a single episode from the receiver's perspective\footnote{The sender is nonetheless able to remember previous time steps from the same episode.}.

Because the receiver is heavily dependent on the messages for the accuracy of its state-value estimates, this mechanism gives the sender very direct feedback about the usefulness of its messages. Under this condition,
the sender mainly models the receiver and only models the environment insofar as it affects the receiver's state-value estimate. Hence, the sender can be said to have an implicit model of the other agent.

\subsection{Experiments}
Agents were trained with the GoToObj environment from the original BabyAI implementation, as well as with custom environments that were derived from it. Figure \ref{fig:types} shows instances of these environments.

The environments contain objects that 
can have one of three types and one of six colors. Unless otherwise noted, object types and colors are sampled randomly.

In each of the environments, the task is for the receiver to move toward an object having the type and color specified in the instruction. The type and color specified correspond to a randomly selected object in the environment. An episode ends once either the receiver faces an object with the specified characteristics or 64 time steps have passed.

The communication interval $n$ was varied logarithmically according to the powers of 2, starting at $n = 1$ and ending at $n = 64$. Since episodes have a maximum length of 64 time steps, $n = 64$ equates to one message per episode. The first message in an episode is always emitted at the first time step of the episode.

The agents' transfer learning abilities were gauged using various pretraining experiments. In these experiments, the agents had to adapt either to additional obstacles, or to increased goal ambiguity.

\paragraph{GoToObj}
This is an environment from the original BabyAI platform. This is an $8 \times 8$ grid, where all cells along the perimeter are intraversible walls, effectively reducing it to a $6 \times 6$ grid. The agent and the object are spawned in different randoms locations on this grid.

\paragraph{GoToObjUnlocked}
This environment was customly designed to complicate the task as seen in the GoToObj environment. Specifically, the agent effectively has to move to two locations in succession instead of one in order to reach the goal. Moreover, in addition to turning and moving forward, the agent must learn to toggle the door to open it. The door connects two $3 \times 3$ rooms in a larger $9 \times 9$ grid. The door is positioned randomly in one of the three cells connecting these rooms. The agent is always spawned in the upper-left room, while the object is always spawned in the upper-right room and is never of the key type.

\setcounter{figure}{2}
\begin{figure*}[bh]
  \centering
  \includegraphics[width=3.35in,trim={0.1in 0.5in 0     0.1in},clip]{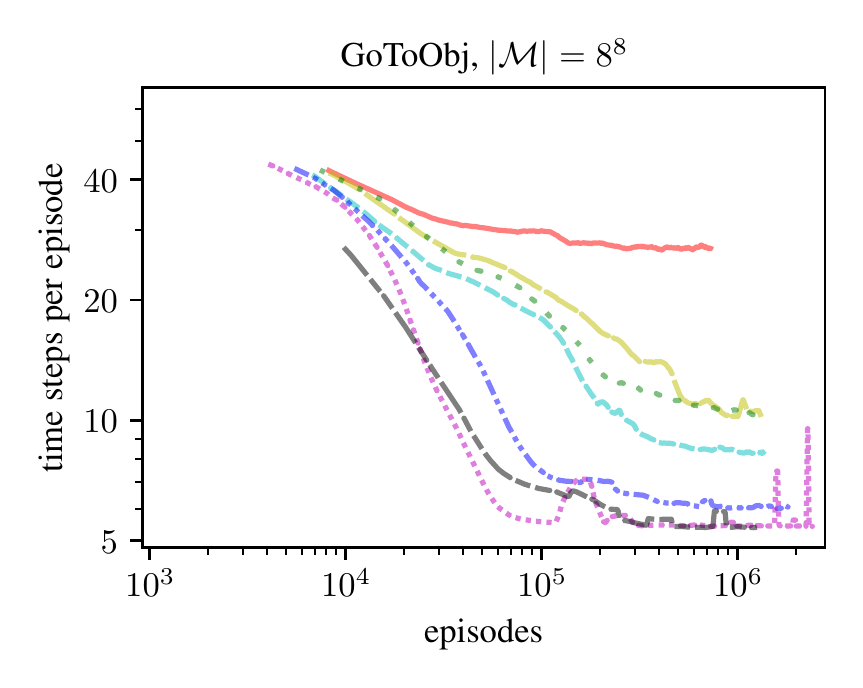}              \includegraphics[width=2.85in,trim={0.5in 0.5in 0.1in 0.1in},clip]{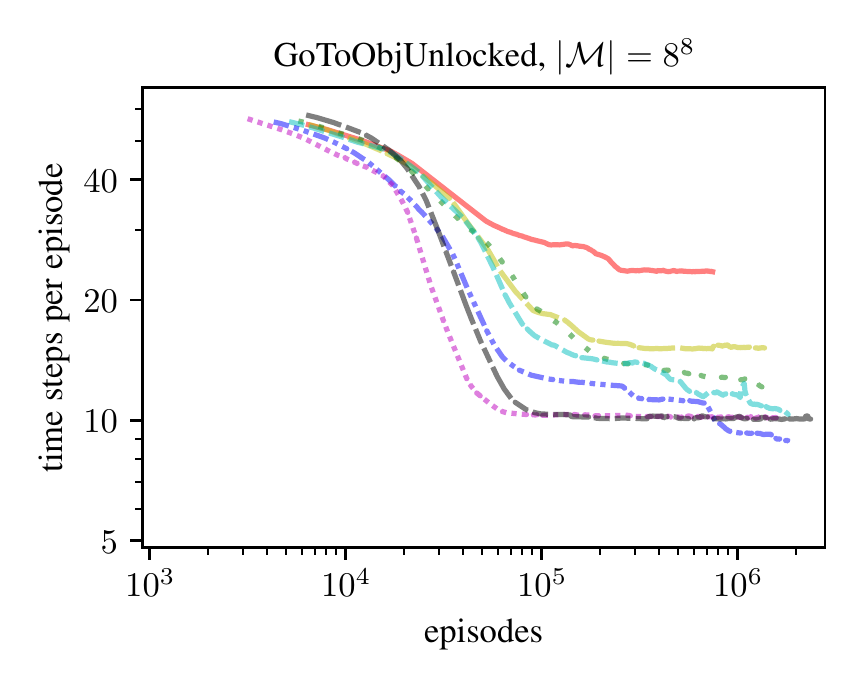}  \\
  \includegraphics[width=3.35in,trim={0.1in 0.1in 0     0    },clip]{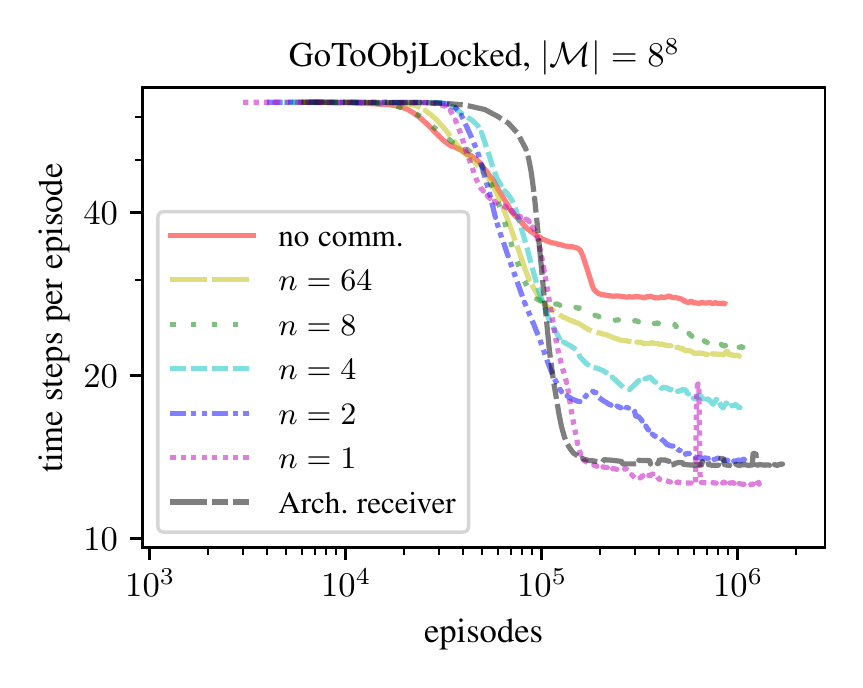} \includegraphics[width=2.85in,trim={0.5in 0.1in 0.1in 0    },clip]{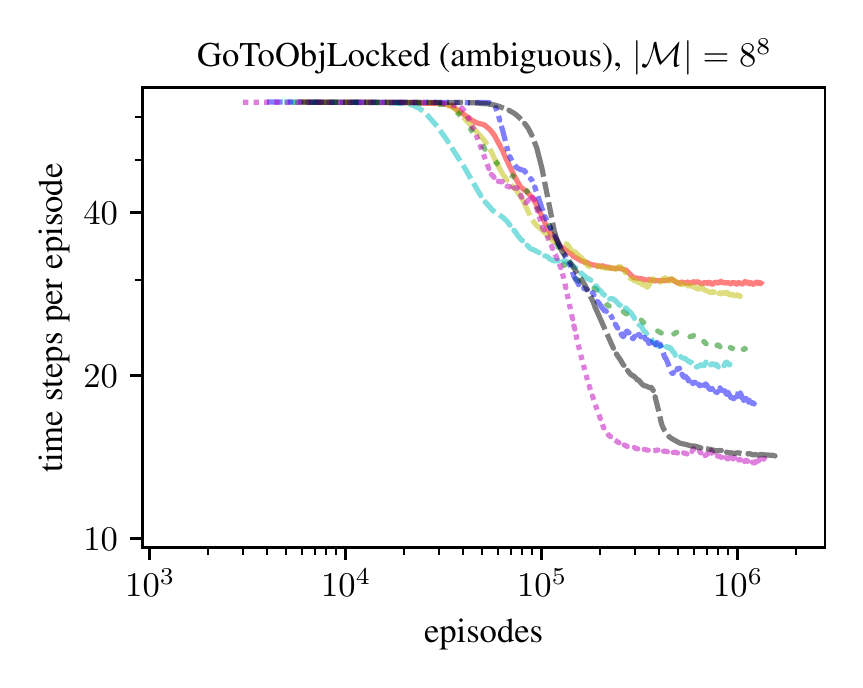}\\
  \caption{log-log plots of learning curves of agents in various environments.}
  \label{fig:results}
\end{figure*}

\paragraph{GoToObjLocked}
This is another customly designed environment that further complicates the task. Not only must the agent learn to turn, move forward and toggle a door, but 
it must learn to pick up a key to unlock the door.

There are two variants of this environment. In the unambiguous variant there is---in addition to the key object---one unambiguous goal object. In the ambiguous variant there is an additional, distracting object. Due to the encyclopedic incentive, the latter variant leaves it ambiguous to the sender which object is the goal. Wherever the variant is not specified, it can be assumed that we are referring to the unambiguous variant.

To ensure that the task can actually be completed, both the agent and the key to the door are spawned in the upper-left room. The goal object and any distracting object are spawned in the upper-right room and are never of the key type.

\section{Results}
The original BabyAI paper used task success as the principal metric to evaluate performance. However, this is not an appropriate metric to assess the benefit of communication in this case; even without communication, the task success rate can reach 100~\%. Rather than in task success, the benefit of communication is primarily in the reduction of the time required for task success. Therefore, we instead measured performance using the number of time steps per episode, i.e., episode length. More precisely, we measured the average over the lengths of all episodes completed in 75 successive batches.
Using this metric, we expected results to be bounded between two extremes. Firstly, disallowing any communication should result in an approximate upper bound on the number of time steps per episode required to complete the task, as the receiver is dependent entirely on its own, very limited observations. Secondly, we might find an approximate lower bound when the receiver has access to all information that the sender has. To obtain a reasonable approximation of the lower bound, we introduced the so-called ``Archimedean\footnote{The moniker ``Archimedean'' was derived from the term ``Archimedean point'' and is meant to be understood as an antonym of ``myopic''.} receiver'' which does not read messages, but is given the sender's observations as a substitute for hypothetical messages that contain all relevant information.

\setcounter{figure}{3}
\begin{figure*}[hb]
  \centering
  \includegraphics[width=3.35in,trim={0.1in 0.1in 0     0    },clip]{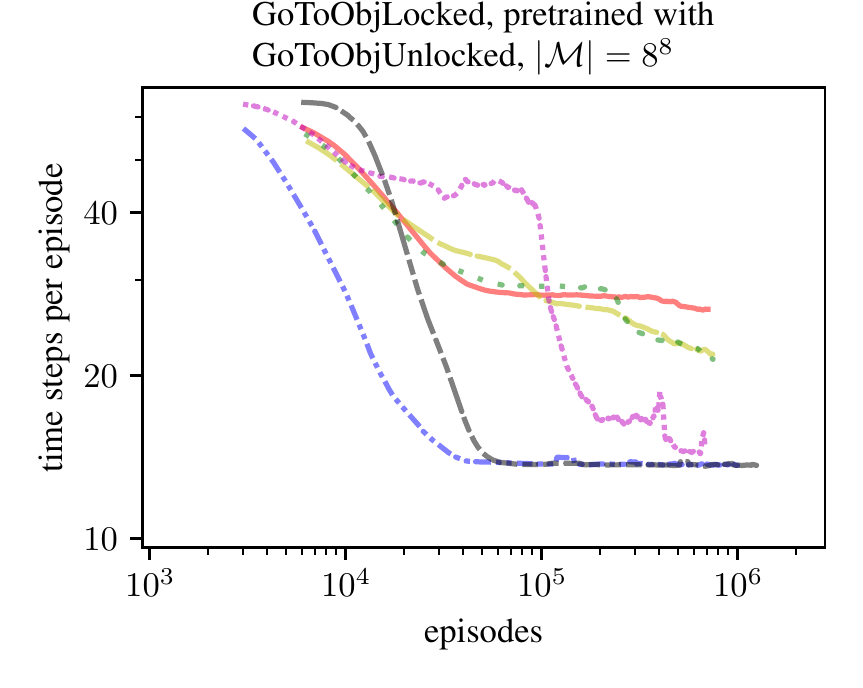}
  \includegraphics[width=2.85in,trim={0.5in 0.1in 0.1in 0    },clip]{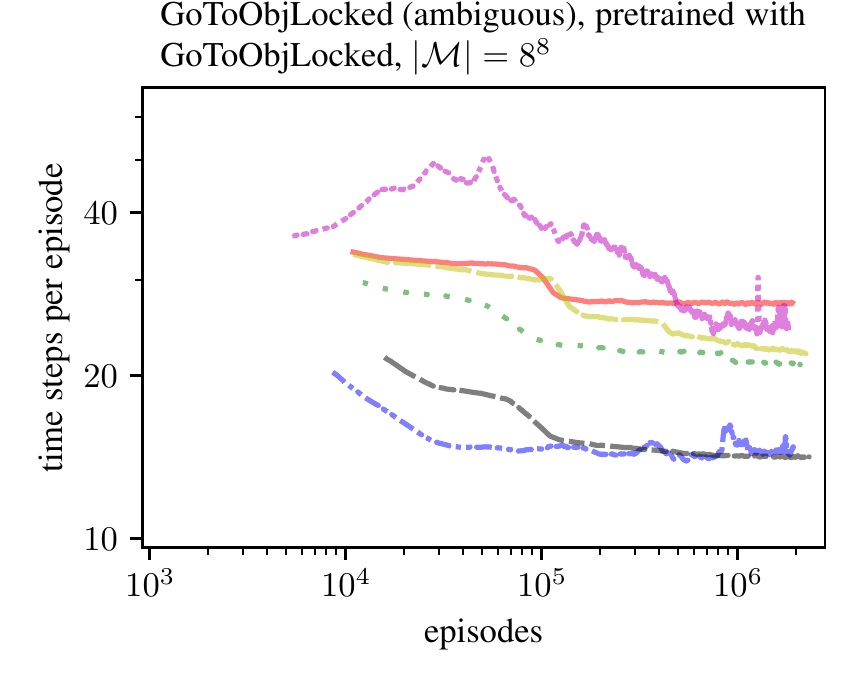}\\
  \caption{log-log plots of learning curves of various pretrained agents.}
  \label{fig:pretraining}
\end{figure*}

With regards to the communication interval, we skipped the values $n = 16$ and $n = 32$ from the powers of 2; while we did perform some experiments with these values, we found these curves to usually fall predictably between those of $n = 8$ and $n = 64$, which already tended to be very similar to each other.

Figure \ref{fig:results} shows the learning curves of agents trained with varying communication intervals in all four environments, while figure \ref{fig:pretraining} shows the learning curves of the variously pretrained agents.

\section{Discussion}
Figure \ref{fig:results} shows a clear separation between the upper and lower bounds in all environments, with the learning curves of agents trained with lower values of $n$ hugging the lower bound and approaching the upper bound for agents trained with increasing values of $n$. This indicates that while communication was of substantial benefit for shorter communication intervals, this benefit decreased with longer intervals. Because the sender relies on the receiver's feedback in the form of its state-value estimates $V_r(S_i)$, we surmise that this decrease is at least in part due to the difficulty for the receiver in learning to make these estimates accurately over such long intervals.

\setcounter{figure}{1}
\begin{figure}[H]
  \centering
  \includegraphics[width=0.85333 in]{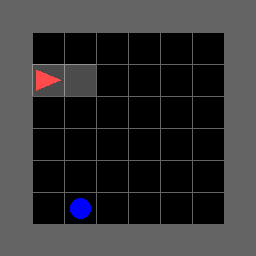} \hspace{0.04 in}
  \includegraphics[width=0.96 in]{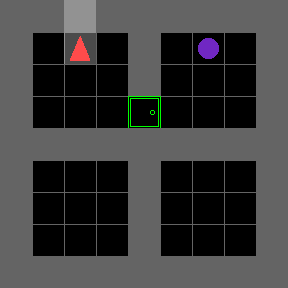} \hspace{0.04 in}
  \includegraphics[width=0.96 in]{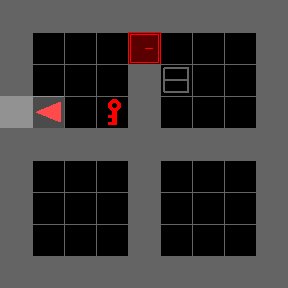}
  \caption{instances of the GoToObj, GoToObjUnlocked and GoToObjLocked environments, from left to right, respectively. The triangle indicates the receiver's location and orientation, while the highlighted cells indicate its partial, egocentric view. In the custom environments, the two connecting rooms are always the two upper rooms. The two bottom rooms are never used and merely serve as padding so that the observations are square, which our implementation of the agent model incidentally requires.
  }
  \label{fig:types}
\end{figure}

Figure \ref{fig:pretraining} shows not only that most agents pretrained with communication were more sample efficient than agents pretrained without communication, but also that some were more sample efficient than the Archimedean receiver. This indicates that emergent communication can be leveraged to improve transfer learning performance.

\subsection{Semantic Analysis}
We analyzed the semantics of the communication protocols that had emerged between agent pairs trained in the GoToObj environment at the end of their respective learning curves in figure \ref{fig:results}. For the purpose of this analysis the messages were not sampled stochastically, but instead determined using the $\arg\,\max$ function.

Table \ref{table:gotoobj_n1} shows the actions taken by the receiver following some of the sender's messages, with $n = 1$. One striking anomaly that in fact holds for all 495 emitted messages $m$ is that none of them have an associated probability $P(a_0 | m) > 0.00$, i.e., the receiver never turned left. We surmise that this suboptimality is idiosyncratic to PPO in general and not to the use of communication specifically. Under the assumption that the receiver would not turn left, there is a clear correlation between table \ref{table:gotoobj_n1} and figure \ref{fig:analysis_bm_n1}; $p(a | m)$ is virtually optimal.

\setcounter{figure}{4}
\begin{figure*}
  \centering
  \includegraphics[height=1.70666667in,trim={0 0 0 0},clip]{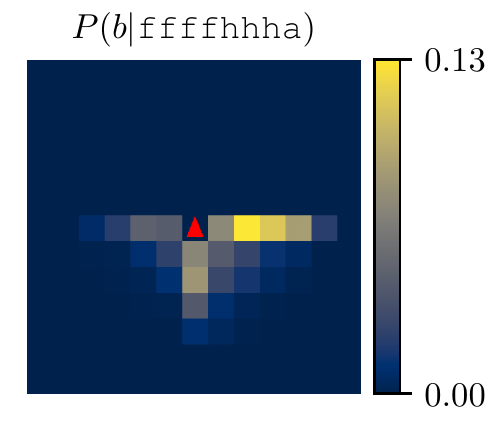}
  \includegraphics[height=1.70666667in,trim={0 0 0 0},clip]{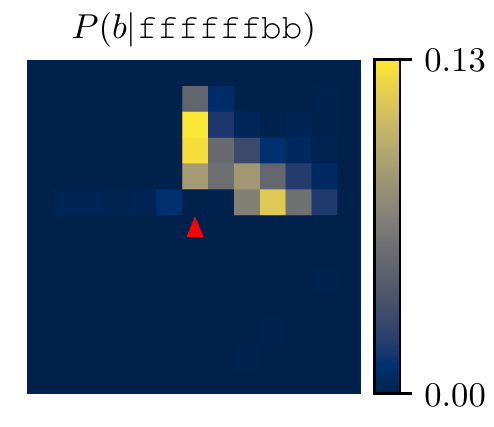}
  \includegraphics[height=1.70666667in,trim={0 0 0 0},clip]{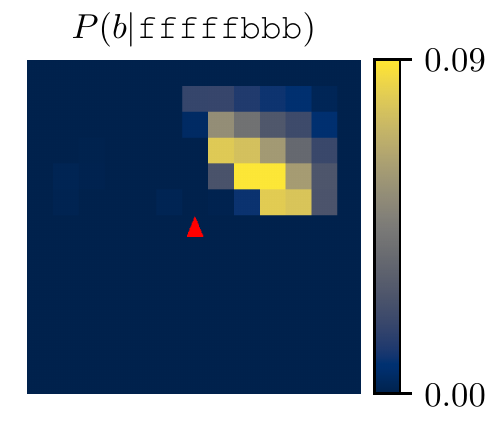}
  \caption{the probabilities $P(b | m)$ that the object was in a specified location $b$ relative to the receiver's location and orientation at a time step in which the sender emitted the message $m$ with $n=1$ in the GoToObj environment. The triangle indicates the receiver's relative location and orientation.}
  \label{fig:analysis_bm_n1}
\end{figure*}

For longer communication intervals, it is not as straightforward to ascertain semantics, as there is a combinatorial explosion of possible trajectories following a message. Table \ref{table:bbbbbbbb:eeeeeeee} shows the actions taken by the receiver averaged over integral trajectories following some of the sender's messages, with $n = 64$. The strategy associated with the most frequent message ``\texttt{bbbbbbbb}'' can be identified as repeatedly moving straight into a wall and then turning left, thereby scouring the perimeter. Consulting figure \ref{fig:analysis_abs_bm_n64}, this appears to be sensible, although it is not optimal behavior. Similarly, the third most frequent message ``\texttt{eeeeeeee}'' is associated with a sensible strategy, if suboptimal.

\section{Conclusion}
We have demonstrated that established deep reinforcement learning techniques are sufficient to incentivize the emergence of grounded discrete communication between agents in grid world environments, without employing straight-through estimation or specialized inductive biases.
In addition, we have shown that such communication can be leveraged to improve transfer learning performance.

\startsquarepar{
One limitation of the incentives that we have provided is the assumption that the receiver supplies a reward signal to the sender. However, this assumption is satisfied naturally by the state-value estimate of any actor--critic agent model and is arguably more straightforward to implement than inductive biases used in previous work. Another limitation is that the effectiveness of communication decreased as the communication interval increased, ostensibly
\unskip\parfillskip 0pt \par}\stopsquarepar

\setlength{\textfloatsep}{0pt plus 0.0pt minus 0.0pt}
\setlength{\intextsep}{0pt plus 0.0pt minus 0.0pt}
\begin{figure}[H]
  \centering
  \includegraphics[width=3in,trim={1.95in 2.65in 1.5in 2.5in},clip]{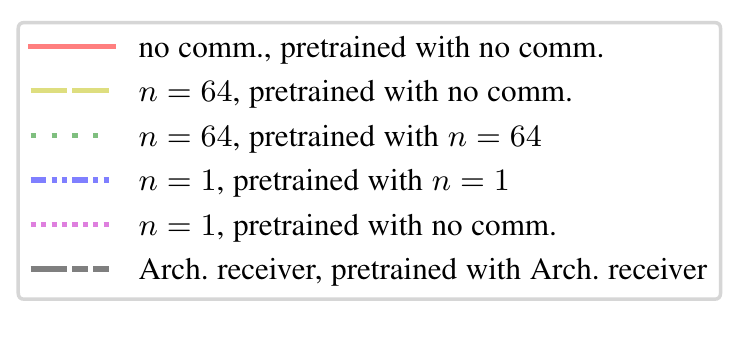}
  \label{fig:pretraining_legend}
\end{figure}
\setlength{\textfloatsep}{20pt plus 2.0pt minus 4.0pt}
\setlength{\intextsep}{12pt plus 2.0pt minus 2.0pt}

\setcounter{table}{1}
\begin{table*}[b]
  \centering
  \setlength\tabcolsep{2pt}
  \begin{tabular}{l @{\hskip 0.57cm} c c r @{\hskip 0.57cm} c c c r}
    $c$      & $P(a_0 | m_0, c)$ & $P(a_2 | m_0, c)$ & $P(m_0 | c)$ & $P(a_0 | m_2, c)$ & $P(a_1 | m_2, c)$ & $P(a_2 | m_2, c)$ & $P(m_2 | c)$ \\ [ 0.5ex]
    \hline                                                                                                                                     \\ [-2.0ex]
    anything &     \textbf{0.22} &     \textbf{0.78} &     21.96~\% &     \textbf{0.24} &     \textbf{0.14} &     \textbf{0.62} &     17.05~\% \\ [ 0.5ex]
    nothing  &              0.00 &     \textbf{1.00} &     16.66~\% &     \textbf{0.12} &     \textbf{0.16} &     \textbf{0.72} &     13.43~\% \\ [ 0.5ex]
    wall     &     \textbf{1.00} &              0.00 &      5.30~\% &     \textbf{1.00} &              0.00 &              0.00 &      3.62~\% \\ [ 0.5ex]
    \hline                                                                                                                                     \\ [-2.0ex]
  \end{tabular}
  \caption{the probability distributions $p(a | m, c)$ of an action $a$ of the receiver following the sender's emission of the most frequent message $m_0 = \texttt{bbbbbbbb}$ or the third most frequent message $m_2 = \texttt{eeeeeeee}$ with $n = 64$ in the GoToObj environment. The distributions are further conditioned on $c$, which here indicates what was in the cell facing the receiver at the time step that the action was taken, namely ``anything'', ``nothing'' or a ``wall''. Actions $a_0$ and $a_1$ are to turn left and right, respectively, while action $a_2$ is to move forward.}
  \label{table:bbbbbbbb:eeeeeeee}
\end{table*}

\noindent as a consequence of a decrease in the accuracy of the reward signal provided by the receiver to the sender.

\subsection{Future Work}
\newcite{kottur2017natural} and \newcite{lowe2019pitfalls} previously stressed that, despite the relative simplicity of emergent communication in contemporary works, the associated semantics are difficult to interpret. Our work ties in to this assessment, as we had difficulty analyzing the semantics in all but the simplest of environments. One sinister future scenario could involve inadvertently allowing agents to use steganography. We therefore consider the development of \textit{artificial language processing} (ALP) to be of crucial importance to understanding artficial intelligence and maintaining value alignment into the far future \cite{yudkowsky2004coherent,leike2018scalable}.

\newcite{DBLP:journals/corr/abs-1203-2990} proposed a theory that relates difficulty of learning in deep architectures to culture and language. Our setup can be used to test this theory, by iteratively training random pairs of agents from a larger pool and observing whether, by chance, a more effective communication protocol emerges between one of the pairs and manages to spread through subsequent pairings.

\setcounter{table}{0}
\begin{table}[H]
  \centering
  \setlength\tabcolsep{5pt}
  \begin{tabular}{l c c r}
    $m$               &  $P(a_1 | m)$ &  $P(a_2 | m)$ &   $P(m)$ \\ [ 0.5ex]
    \hline                                                       \\ [-2.0ex]
    \texttt{ffffhhha} & \textbf{1.00} &          0.00 & 11.67~\% \\ [ 0.5ex]
    \texttt{ffffffbb} &          0.00 & \textbf{1.00} &  9.10~\% \\ [ 0.5ex]
    \texttt{fffffbbb} &          0.00 & \textbf{1.00} &  8.50~\% \\ [ 0.5ex]
    \hline                                                       \\ [-2.0ex]
    \multicolumn{4}{r}{total: 29.27~\%}
  \end{tabular}
  \caption{the probability distributions $p(a | m)$ of the first action $a$ of the receiver following the sender's emission of one of the three most frequent messages $m$ with $n = 1$ in the GoToObj environment. Action $a_1$ is to turn right, while action $a_2$ is to move forward.}
  \label{table:gotoobj_n1}
\end{table}

\vfill

\setcounter{figure}{5}
\begin{figure}[H]
  \centering
  \includegraphics[height=1.17333333in]{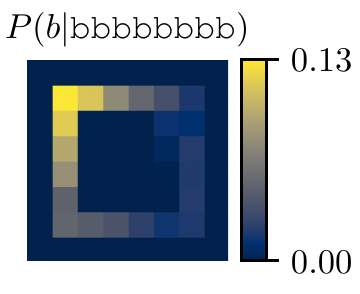}
  \includegraphics[height=1.17333333in]{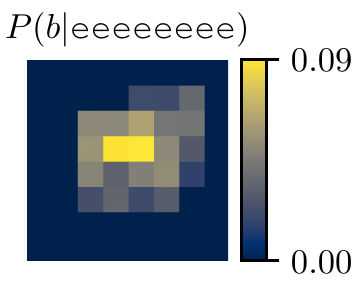}
  \caption{the probabilities $P(b | m)$ that the object was in a specified location $b$ relative to the receiver's orientation at a time step in which the sender emitted the message $m$ with $n=64$ in the GoToObj environment.}
  \label{fig:analysis_abs_bm_n64}
\end{figure}

\newpage

\section*{Acknowledgments}
We thank Tim Baumg\"artner, Gautier Dagan, Wilker Ferreira Aziz, Dieuwke Hupkes, Bence Keresztury, Mathijs Mul, Diana Rodr\'iguez Luna and Sainbayar Sukhbaatar for offering their help in producing this work.

\bibliography{emnlp2020}

\begin{thebibliography}{23}
\expandafter\ifx\csname natexlab\endcsname\relax\def\natexlab#1{#1}\fi

\bibitem[{Bengio(2012)}]{DBLP:journals/corr/abs-1203-2990}
Yoshua Bengio. 2012.
\newblock \href {http://arxiv.org/abs/1203.2990} {{E}volving {C}ulture vs
  {L}ocal {M}inima}.
\newblock \emph{CoRR}, abs/1203.2990.

\bibitem[{Bengio et~al.(2013)Bengio, L{\'{e}}onard, and
  Courville}]{DBLP:journals/corr/BengioLC13}
Yoshua Bengio, Nicholas L{\'{e}}onard, and Aaron~C. Courville. 2013.
\newblock \href {http://arxiv.org/abs/1308.3432} {{E}stimating or {P}ropagating
  {G}radients {T}hrough {S}tochastic {N}eurons for {C}onditional
  {C}omputation}.
\newblock \emph{CoRR}, abs/1308.3432.

\bibitem[{Chevalier{-}Boisvert et~al.(2018)Chevalier{-}Boisvert, Bahdanau,
  Lahlou, Willems, Saharia, Nguyen, and
  Bengio}]{DBLP:journals/corr/abs-1810-08272}
Maxime Chevalier{-}Boisvert, Dzmitry Bahdanau, Salem Lahlou, Lucas Willems,
  Chitwan Saharia, Thien~Huu Nguyen, and Yoshua Bengio. 2018.
\newblock \href {http://arxiv.org/abs/1810.08272} {Baby{AI}: First {S}teps
  {T}owards {G}rounded {L}anguage {L}earning with a {H}uman in the {L}oop}.
\newblock \emph{CoRR}, abs/1810.08272.

\bibitem[{Cho et~al.(2014)Cho, Van~Merri{\"e}nboer, Gulcehre, Bahdanau,
  Bougares, Schwenk, and Bengio}]{cho2014learning}
Kyunghyun Cho, Bart Van~Merri{\"e}nboer, Caglar Gulcehre, Dzmitry Bahdanau,
  Fethi Bougares, Holger Schwenk, and Yoshua Bengio. 2014.
\newblock Learning {P}hrase {R}epresentations {U}sing {RNN} {E}ncoder-{D}ecoder
  for {S}tatistical {M}achine {T}ranslation.
\newblock \emph{arXiv preprint arXiv:1406.1078}.

\bibitem[{Choi et~al.(2018)Choi, Lazaridou, and
  de~Freitas}]{choi2018compositional}
Edward Choi, Angeliki Lazaridou, and Nando de~Freitas. 2018.
\newblock Compositional {O}bverter {C}ommunication {L}earning from {R}aw
  {V}isual {I}nput.
\newblock \emph{arXiv preprint arXiv:1804.02341}.

\bibitem[{Das et~al.(2018)Das, Gervet, Romoff, Batra, Parikh, Rabbat, and
  Pineau}]{DBLP:journals/corr/abs-1810-11187}
Abhishek Das, Th{\'{e}}ophile Gervet, Joshua Romoff, Dhruv Batra, Devi Parikh,
  Michael Rabbat, and Joelle Pineau. 2018.
\newblock \href {http://arxiv.org/abs/1810.11187} {Tar{MAC}: Targeted
  {M}ulti-{A}gent {C}ommunication}.
\newblock \emph{CoRR}, abs/1810.11187.

\bibitem[{Foerster et~al.(2016)Foerster, Assael, de~Freitas, and
  Whiteson}]{foerster2016learning}
Jakob Foerster, Ioannis~Alexandros Assael, Nando de~Freitas, and Shimon
  Whiteson. 2016.
\newblock Learning to {C}ommunicate with {D}eep {M}ulti-{A}gent {R}einforcement
  {L}earning.
\newblock In \emph{Advances in Neural Information Processing Systems}, pages
  2137--2145.

\bibitem[{Havrylov and Titov(2017)}]{DBLP:journals/corr/HavrylovT17}
Serhii Havrylov and Ivan Titov. 2017.
\newblock \href {http://arxiv.org/abs/1705.11192} {Emergence of {L}anguage with
  {M}ulti-{A}gent {G}ames: {L}earning to {C}ommunicate with {S}equences of
  {S}ymbols}.
\newblock \emph{CoRR}, abs/1705.11192.

\bibitem[{Hernandez{-}Leal et~al.(2018)Hernandez{-}Leal, Kartal, and
  Taylor}]{DBLP:journals/corr/abs-1810-05587}
Pablo Hernandez{-}Leal, Bilal Kartal, and Matthew~E. Taylor. 2018.
\newblock \href {http://arxiv.org/abs/1810.05587} {Is {M}ulti-{A}gent {D}eep
  {R}einforcement {L}earning the {A}nswer or the {Q}uestion? {A} {B}rief
  {S}urvey}.
\newblock \emph{CoRR}, abs/1810.05587.

\bibitem[{Hochreiter and Schmidhuber(1997)}]{hochreiter1997long}
Sepp Hochreiter and J{\"u}rgen Schmidhuber. 1997.
\newblock Long {S}hort-{T}erm {M}emory.
\newblock \emph{Neural Computation}, 9(8):1735--1780.

\bibitem[{Ioffe and Szegedy(2015)}]{ioffe2015batch}
Sergey Ioffe and Christian Szegedy. 2015.
\newblock {Batch Normalization: Accelerating Deep Network Training by Reducing
  Internal Covariate Shift}.
\newblock In \emph{International Conference on Machine Learning}, pages
  448--456.

\bibitem[{Jang et~al.(2016)Jang, Gu, and Poole}]{jang2016categorical}
Eric Jang, Shixiang Gu, and Ben Poole. 2016.
\newblock {C}ategorical {R}eparameterization with {G}umbel-{S}oftmax.
\newblock \emph{arXiv preprint arXiv:1611.01144}.

\bibitem[{Jaques et~al.(2018)Jaques, Lazaridou, Hughes, G{\"{u}}l{\c{c}}ehre,
  Ortega, Strouse, Leibo, and de~Freitas}]{DBLP:journals/corr/abs-1810-08647}
Natasha Jaques, Angeliki Lazaridou, Edward Hughes, {\c{C}}aglar
  G{\"{u}}l{\c{c}}ehre, Pedro~A. Ortega, DJ~Strouse, Joel~Z. Leibo, and Nando
  de~Freitas. 2018.
\newblock \href {http://arxiv.org/abs/1810.08647} {Intrinsic {S}ocial
  {M}otivation {V}ia {C}ausal {I}nfluence in {M}ulti-{A}gent {RL}}.
\newblock \emph{CoRR}, abs/1810.08647.

\bibitem[{Kottur et~al.(2017)Kottur, Moura, Lee, and Batra}]{kottur2017natural}
Satwik Kottur, Jos{\'e} Moura, Stefan Lee, and Dhruv Batra. 2017.
\newblock {Natural Language Does Not Emerge `Naturally' in Multi-Agent Dialog}.
\newblock In \emph{Proceedings of the 2017 Conference on Empirical Methods in
  Natural Language Processing}, pages 2962--2967.

\bibitem[{Lazaridou et~al.(2018)Lazaridou, Hermann, Tuyls, and
  Clark}]{DBLP:journals/corr/abs-1804-03984}
Angeliki Lazaridou, Karl~Moritz Hermann, Karl Tuyls, and Stephen Clark. 2018.
\newblock \href {http://arxiv.org/abs/1804.03984} {Emergence of {L}inguistic
  {C}ommunication from {R}eferential {G}ames with {S}ymbolic and {P}ixel
  {I}nput}.
\newblock \emph{CoRR}, abs/1804.03984.

\bibitem[{Leike et~al.(2018)Leike, Krueger, Everitt, Martic, Maini, and
  Legg}]{leike2018scalable}
Jan Leike, David Krueger, Tom Everitt, Miljan Martic, Vishal Maini, and Shane
  Legg. 2018.
\newblock Scalable {A}gent {A}lignment {V}ia {R}eward {M}odeling: {A}
  {R}esearch {D}irection.
\newblock \emph{arXiv preprint arXiv:1811.07871}.

\bibitem[{Lowe et~al.(2019)Lowe, Foerster, Boureau, Pineau, and
  Dauphin}]{lowe2019pitfalls}
Ryan Lowe, Jakob Foerster, Y-Lan Boureau, Joelle Pineau, and Yann Dauphin.
  2019.
\newblock {On the Pitfalls of Measuring Emergent Communication}.
\newblock In \emph{Proceedings of the 18th International Conference on
  Autonomous Agents and MultiAgent Systems}, pages 693--701. International
  Foundation for Autonomous Agents and Multiagent Systems.

\bibitem[{Maddison et~al.(2016)Maddison, Mnih, and
  Teh}]{DBLP:journals/corr/MaddisonMT16}
Chris~J. Maddison, Andriy Mnih, and Yee~Whye Teh. 2016.
\newblock \href {http://arxiv.org/abs/1611.00712} {{T}he {C}oncrete
  {D}istribution: {A} {C}ontinuous {R}elaxation of {D}iscrete {R}andom
  {V}ariables}.
\newblock \emph{CoRR}, abs/1611.00712.

\bibitem[{Mordatch and Abbeel(2018)}]{mordatch2018emergence}
Igor Mordatch and Pieter Abbeel. 2018.
\newblock Emergence of {G}rounded {C}ompositional {L}anguage in {M}ulti-{A}gent
  {P}opulations.
\newblock In \emph{Thirty-{S}econd {AAAI} {C}onference on {A}rtificial
  {I}ntelligence}.

\bibitem[{Perez et~al.(2018)Perez, Strub, De~Vries, Dumoulin, and
  Courville}]{perez2018film}
Ethan Perez, Florian Strub, Harm De~Vries, Vincent Dumoulin, and Aaron
  Courville. 2018.
\newblock Fi{LM}: {V}isual {R}easoning with a {G}eneral {C}onditioning {L}ayer.
\newblock In \emph{Thirty-Second AAAI Conference on Artificial Intelligence}.

\bibitem[{Radford et~al.(2019)Radford, Wu, Child, Luan, Amodei, and
  Sutskever}]{radford2019language}
Alec Radford, Jeff Wu, Rewon Child, David Luan, Dario Amodei, and Ilya
  Sutskever. 2019.
\newblock Language {M}odels {A}re {U}nsupervised {M}ultitask {L}earners.

\bibitem[{Schulman et~al.(2017)Schulman, Wolski, Dhariwal, Radford, and
  Klimov}]{schulman2017proximal}
John Schulman, Filip Wolski, Prafulla Dhariwal, Alec Radford, and Oleg Klimov.
  2017.
\newblock Proximal {P}olicy {O}ptimization {A}lgorithms.
\newblock \emph{arXiv preprint arXiv:1707.06347}.

\bibitem[{Yudkowsky(2004)}]{yudkowsky2004coherent}
Eliezer Yudkowsky. 2004.
\newblock Coherent {E}xtrapolated {V}olition.
\newblock \emph{Singularity Institute for Artificial Intelligence}.

\end{thebibliography}
\bibliographystyle{acl_natbib}

\end{document}